\documentclass{svproc}
\usepackage{graphicx}%
\usepackage{multirow}%
\usepackage{mathrsfs}%
\usepackage[title]{appendix}%
\usepackage{xcolor}%
\usepackage{textcomp}%
\usepackage{manyfoot}%
\usepackage{booktabs}%
\usepackage{algorithm}%
\usepackage{algorithmicx}%
\usepackage{algpseudocode}%
\usepackage{listings}%
\usepackage{url}
\usepackage[caption=false,font=footnotesize]{subfig}

\usepackage{float}
\usepackage{bm}
\usepackage{enumerate}
\usepackage{color, colortbl}
\usepackage{booktabs, multirow}
\definecolor{darkgreen}{rgb}{0, 0.5, 0}
\definecolor{red}{rgb}{1, 0, 0}
\definecolor{purple}{rgb}{0.5, 0, 0.5}
\usepackage{chngpage}
\usepackage{comment}
\usepackage{mfirstuc}
\usepackage{mathtools}
\usepackage{lettrine}
\usepackage{pxfonts}

\newcommand\ie{\textit{i.e.}}
\newcommand\eg{\textit{e.g.,}}

\newcommand{\norm}[1]{\left\lVert#1\right\rVert}
\newcommand{\Rbb}{{\mathbb{R}}}
\providecommand{\abs}[1]{\left| #1 \right|}
\newcommand{\beq}{\begin{equation}}
\newcommand{\eeq}{\end{equation}}

\newcommand{\diag}{\mathrm{diag}}

\newcommand{\Y}{{\boldsymbol{Y}}}

\renewcommand{\u}{\boldsymbol{u}}

\newcommand{\x}{{\boldsymbol{x}}}

\begin{document}
\mainmatter              
\title{Training Matters: Unlocking Potentials of Deeper Graph Convolutional Neural Networks}
\titlerunning{Unlocking Potentials of Deeper Graph Convolutional Neural Networks}  
%
\author{
Sitao Luan$^{1,2}$, Mingde Zhao$^{1,2}$, Xiao-Wen Chang$^{1}$, Doina Precup$^{1,2,3}$\\
\{sitao.luan@mail, mingde.zhao@mail, chang@cs, dprecup@cs\}.mcgill.ca\\
$^1$McGill University; $^2$Mila; $^3$DeepMind
}

\institute{}
\authorrunning{Sitao Luan et al.} 

\maketitle              

\begin{abstract}
The performance limit of deep Graph Convolutional Networks (GCNs) are pervasively thought to be caused by the inherent limitations of the GCN layers, such as their \textit{insufficient expressive power}. However, if this were true, modifying only the training procedure for a given architecture would not likely to enhance performance. Contrary to this belief, our paper demonstrates several ways to achieve such improvements. We begin by highlighting the training challenges of GCNs from the perspective of graph signal energy loss. More specifically, we find that the loss of energy in the backward pass during training hinders the learning of the layers closer to the input. To address this, we propose several strategies to mitigate the training problem by slightly modifying the GCN operator, from the energy perspective. After empirical validation, we confirm that these changes of operator lead to significant decrease in the training difficulties and notable performance boost, without changing the composition of parameters. With these, we conclude that the root cause of the problem is more likely the \textit{training difficulty} than the others.
\end{abstract}

\section{Introduction}

As a structure that is capable of modeling relational information \cite{hamilton2017inductive,kipf2016classification,gilmer2017neural,monti2017geometric,defferrard2016fast}, graph has inspired the emerge of Graph Neural Networks (GNNs), a machine learning paradigm that achieve state-of-the-art performance on complex tasks \cite{shuman2012emerging,bronstein2016geometric,defferrard2016fast,kipf2016classification,chen2018fastgcn,chen2017stochastic,liao2019lanczos,lim2021large,luan2021heterophily,luan2022complete,luan2022revisiting,luan2023graph,luan2022we}.

GCN \cite{kipf2016classification}, being arguably the most popular method of all GNNs, is applied pervasively for being lightweight and having relatively capable performance. However, the development of GCNs on more complicated tasks is hindered by the fact that their performance is still relatively limited and cannot be easily boosted: the capacity of GCN seems not scalable with the depth of the architectures, while the performance of typical deep learning architectures mostly becomes better with the increment of the depth. Several investigations about the possible cause of the problem have been carried out, including
\begin{itemize}
\item
Oversmoothing Problem \cite{li2018deeper}: stacking aggregation operations in GNNs is shown to make the representation of connected nodes to be more indistinguishable and therefore causes information loss;
\item
Loss of rank \cite{luan2019break}: the numerical ranks of the outputs in hidden layers will decrease with the increment of network depth.
\item
Inevitable convergence to some subspace \cite{oono2019graph}: the layer outputs get closer to a fixed subspace with the increment of the network depth;
\end{itemize}

These analyses show that despite the increment of trainable parameters, simply deepening GCNs is not helpful, therefore it is more promising to just switch to alternate solutions. Following these, efforts have been made to propose alternate GCN architectures to increase the expressive power with additional computational expenses, \eg{} augmenting architectures with layer concatenation operations  \cite{he2016deep,luan2019break}. However, the computational costs introduced often outweigh the performance boost, therefore no alternative is yet popular enough to replace GCN.

The intractability of deep GCNs naturally leads to the belief that deeper GCNs \textit{cannot be trained well} and \textit{cannot have better performance} without the change of architectures. However, in this paper, we question such idea and argue that the crucial factor limiting the performance of GCN architectures is more likely to be the \textit{difficulty in training} instead of \textit{insufficient expressive power}. First, from graph signal energy perspective, we prove that, during training, the energy loss during backward pass makes the training of layers that are closer to the input difficult. Then, we show both in theory and in experiments, it is actually possible, in several ways, to significantly lower training difficulty and gain notable performance boost by only changing slightly the training process of deep GCN, without changing the expressive power. These observations lead us to the discovery that the performance limit of GCN is more likely to be caused by inappropriate training rather than GCNs being inherently incapable.

The methodologies we propose in this paper includes Topology Rescaling (TR) for the graph operator (\eg{} graph Laplacian), weight normalization, energy normalization and weight initialization for enhancing the training of the parameters in the layers, as well as skip (residual) connections that do not use concatenation of layer outputs, \ie{} no additional parameters.

The paper is organized as follows. In Section \ref{sec:preliminary}, we introduce backgrounds of graph Laplacian, graph partition and graph signal energy. In Section \ref{sec:energy_gradient_analysis}, we analyze from the perspective of energy and gradient, arguing that \textit{the energy loss of the backward pass during training} leads to training difficulty, which we will in the end verify as the core factor limiting the performance of GNNs. In Section \ref{sec:alleviate_methods}, we propose $4$ methodologies that addresses the training difficulty problem from different perspectives. In Section \ref{sec:experiments}, we validate the effectiveness of the methods in lowering the training difficulty.

\section{Preliminaries} \label{sec:preliminary}

We use bold fonts for vectors (\eg{} $\bm{v}$), block vectors (\eg{} $\bm{V}$) and matrix blocks (\eg{} $\bm{V_i}$). Suppose we have an undirected connected graph $\mathcal{G}=(\mathcal{V},\mathcal{E})$ without a bipartite component, where $\mathcal{V}$ is the node set with $\abs{\mathcal{V}}=N$, $\mathcal{E}$ is the edge set with $\abs{\mathcal{E}}=E$.
Let $A \in \Rbb^{N\times N}$ be the adjacency matrix of $\mathcal{G}$, \ie{} $A_{ij} = 1$ for $e_{ij} \in \mathcal{E}$ and $A_{ij}=0$ otherwise.
The graph Laplacian is defined as $L=D-A$, where $D$ is a diagonal degree matrix with $D_{ii} = \sum_j A_{ij}$. 
The symmetric normalized Laplacian is defined as $L_{\text{sym}}=I-D^{-1/2} A D^{-1/2}$ with eigenvalues $\lambda(L_{\text{sym}}) \in [0,2)$ and its renormalized version is defined as 
\beq \label{eq:tlsym}
\tilde{L}_{\text{sym}}=I-\tilde{D}^{-1/2} \tilde{A}\tilde{D}^{-1/2}, \ \ 
\tilde{A} = A+I,\ \ \tilde{D}=\diag(\tilde{D}_{ii}), \ \  \textstyle{\tilde{D}_{ii} = \sum_j \tilde{A}_{ij}}
\eeq
and its eigenvalues $\lambda(\tilde{L}_{\text{sym}}) \in [0,2)$ \cite{chung1997spectral}.

The eigendecomposition of $L$ gives us $L=U \Lambda U^{-1}$, where $U=[\u_1,\ldots,\u_N]\in \Rbb^{N\times N}$ is formed by the orthonormal 
eigenvectors, referred to as the \textit{graph Fourier basis}, and $\Lambda=\diag(\lambda_1,\ldots,\lambda_N)$ is formed by the eigenvalues,
which are nonnegative and are referred to as \textit{frequencies}.
Traditionally, graph Fourier basis is defined specifically by eigenvectors of $L$, but in this paper, graph Fourier basis is formed by eigenvectors of the Laplacian we use. The smaller eigenvalue $\lambda_i$ indicates larger global smoothness of $\bm{u}_i$ \cite{dakovic2019local}, which means any two elements of $\bm{u_i}$ corresponding to two directly connected nodes will have similar values. Thus, $\bm{u_i}$ with small $\lambda_i$ tends to partition the graph into large communities. This property is crucial for later analysis.

A graph signal is a vector $\bm{x} \in \mathbb{R}^N$ defined on $\mathcal{V}$, where $x_i$ is defined on the node $i$. We also have a feature matrix (graph signals) $\bm{X} \in \mathbb{R}^{N\times F}$ whose columns are graph signals and each node $i$ has a feature vector $\bm{X_{i,:}}$, which is the $i$-th row of $\bm{X}$. 
The graph Fourier transform of the graph signal $\x$ is defined as $\bm{x}_\mathcal{F} = U^{-1} \bm{x} = U^{T} \bm{x} = [\u_1^T\x, \ldots, \u_N^T\x]^T$, where $\bm{u}_i^T \bm{x}$ is the component of $\bm{x}$ in the direction of $\bm{u_i}$.

In addition to various graph Laplacians, various affinity matrices derived from graph Laplacians 
have been adopted in GNNs. 
The most widely used one is the renormalized affinity matrix
$$ 
\hat{A} \equiv I- \tilde{L}_{sym} = \tilde{D}^{-1/2} \tilde{A} \tilde{D}^{-1/2}
$$
with $\lambda(\hat{A}) = 1-\lambda(\tilde{L}_{sym}) \in (-1,1]$, 
and it is used in GCN \cite{kipf2016classification} as follows
\begin{equation}
    \label{eq:gcn_original}
   \bm{Y} = \text{softmax} (\hat{A} \; \text{ReLU} (\hat{A} \bm{X} W_0 ) \; W_1 )
\end{equation}
where $W_0 \in \Rbb^{F\times F_1}$ and $W_1 \in \Rbb^{F_1\times O}$ are parameter matrices.

\begin{definition}(Energy of signal on graph \cite{gavili2017shift,stankovic2018reduced})
For a signal $\bm{x}$ defined on graph $\mathcal{G}$, its energy is defined as $\norm{\x_{\mathcal{F}}}_2^2$, where $x_{\mathcal{F}}$ is the graph Fourier transform of $\bm{x}$. 
\end{definition}

The energy represents the intensity of a graph signal projected onto the frequency domain. However, considering undirected graph ${\cal G}$, the graph Laplacian is symmetric and the graph Fourier basis matrix is orthogonal, leading to  $\|\x_{\cal F}\|_2^2 = \|\x\|_2^2=\sum_i {x_i}^2$,
which depends on only the signal itself.

\begin{definition}(Energy-preserving operator \cite{gavili2017shift}
or isometric operator\cite{girault2015translation})
An operator $\Phi$ defined on graph signal is \textit{energy-preserving} if for any graph signal $\bm{x}$, it satisfies $\norm{(\Phi \bm{x})_{\mathcal{F}}}_2^2 = \norm{ \bm{x}_{\mathcal{F}}}_2^2$.
\end{definition}

The energy-preserving property means the operator does not change the energy intensity in the frequency domain after being applied on graph signals.

\section{Energy Loss during Back Propagation} 
\label{sec:energy_gradient_analysis}

In this section, we first show that $\text{ReLU}(\hat{A} \cdot)$ is an \textit{energy-losing} operator. This property is the natural explanation for the over-smoothing \cite{li2018deeper}, loss of rank\cite{luan2019break} and loss of expressive power \cite{oono2019graph} phenomena, from which deep GCN will suffer during feed-forward process. According to the above analysis, the top layers will lose signal energy more serious than bottom layers. However, we will show that, contrary to our empirical intuition, deep GCNs lose energy in bottom layers instead of in top layers. Rather than investigating  from feed-forward perspective, we will explain this contradiction from backward view by analyzing the gradient propagation in the following section.

\subsection{Forward Pass Analyses:  Difficult and Complicated}

\begin{theorem} 1 \label{thm:1}
For any undirected connected graph $\mathcal{G}$, 
$\text{ReLU}(\hat{A} \cdot)$ is an energy-losing graph operator, i.e., 
for any graph signal  $\bm{x}$ 
$$
\norm{\left(\text{ReLU}(\hat{A} \bm{x}) \right)_\mathcal{F}}_2^2 \leq \norm{\bm{x}_\mathcal{F}}_2^2
$$
The strict inequality holds for any $\x$ which is independent of 
$[\tilde{D}_{11}^{1/2}, \ldots, \tilde{D}_{NN}^{1/2}]^T$,
where $\tilde{D}_{ii}$ for $i=1,\ldots,N$ are defined in \eqref{eq:tlsym}
\end{theorem}

\begin{figure*}[htbp]
\centering
\subfloat[]{
\captionsetup{justification = centering}
\includegraphics[width=0.32\textwidth]{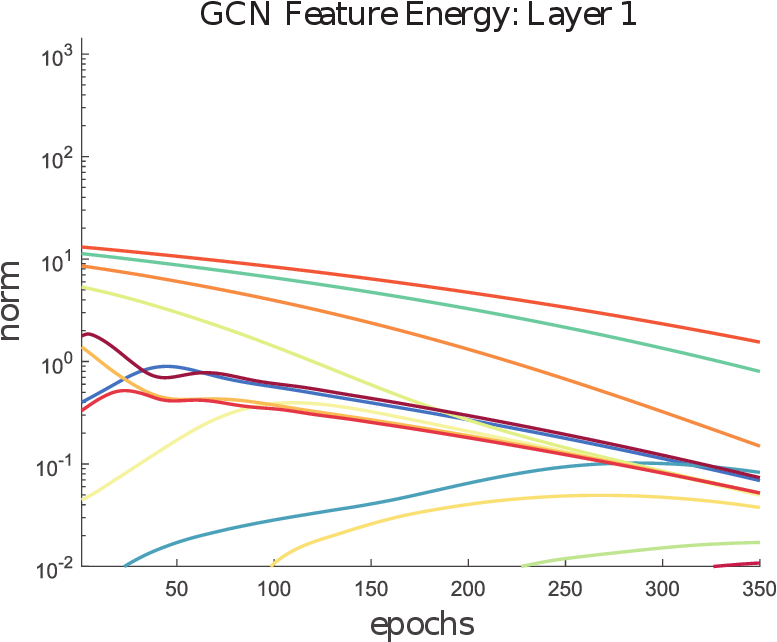}}
\hfill
\subfloat[]{
\captionsetup{justification = centering}
\includegraphics[width=0.32\textwidth]{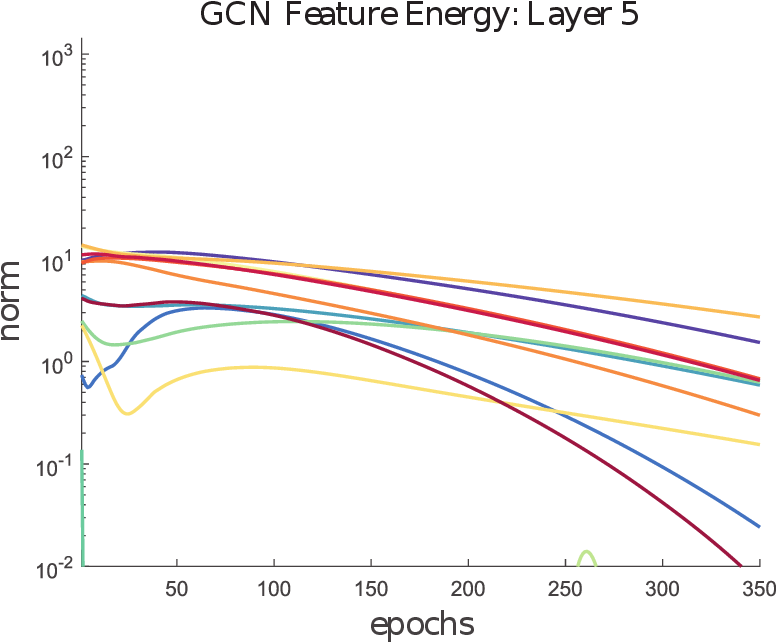}}
\hfill
\subfloat[]{
\captionsetup{justification = centering}
\includegraphics[width=0.32\textwidth]{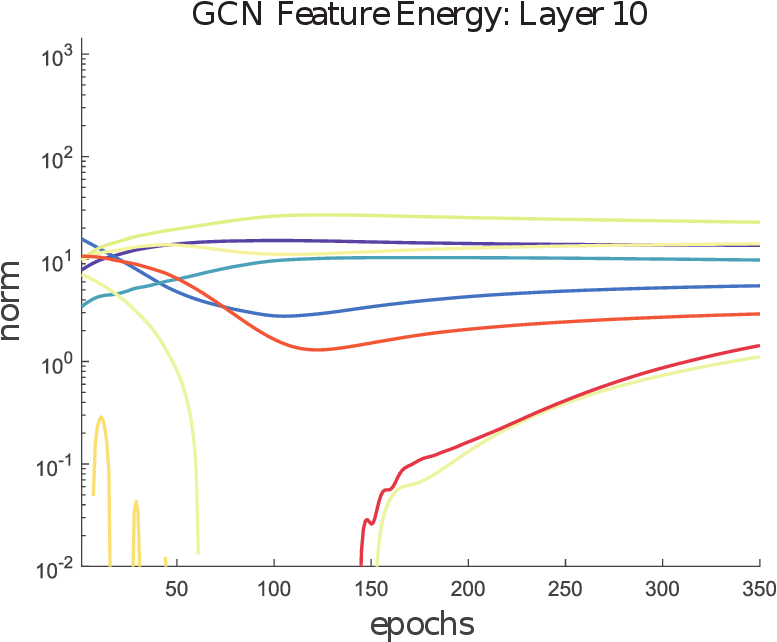}}
\vfill
\subfloat[]{
\captionsetup{justification = centering}
\includegraphics[width=0.32\textwidth]{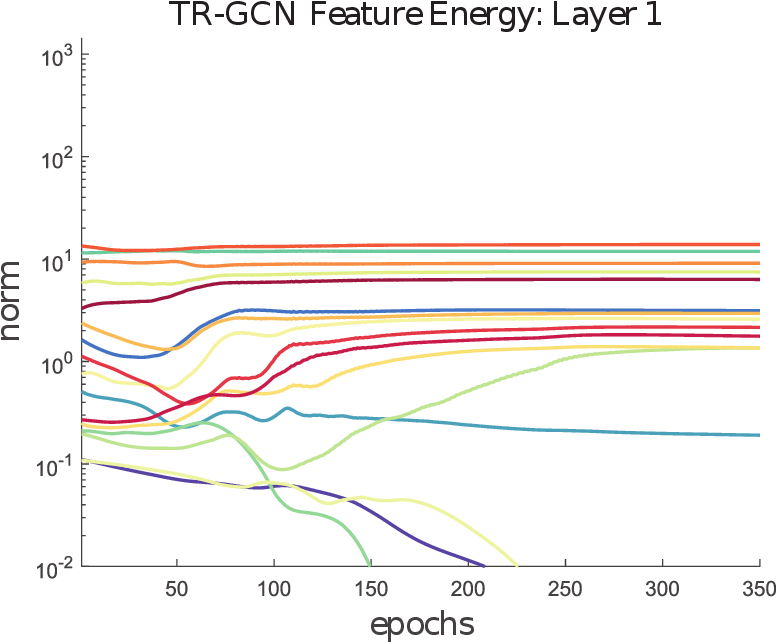}}
\hfill
\subfloat[]{
\captionsetup{justification = centering}
\includegraphics[width=0.32\textwidth]{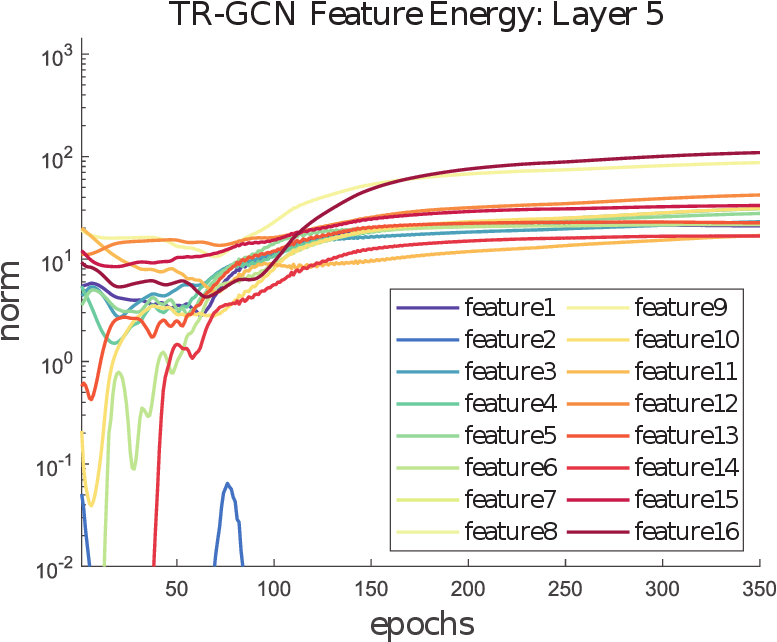}}
\hfill
\subfloat[]{
\captionsetup{justification = centering}
\includegraphics[width=0.32\textwidth]{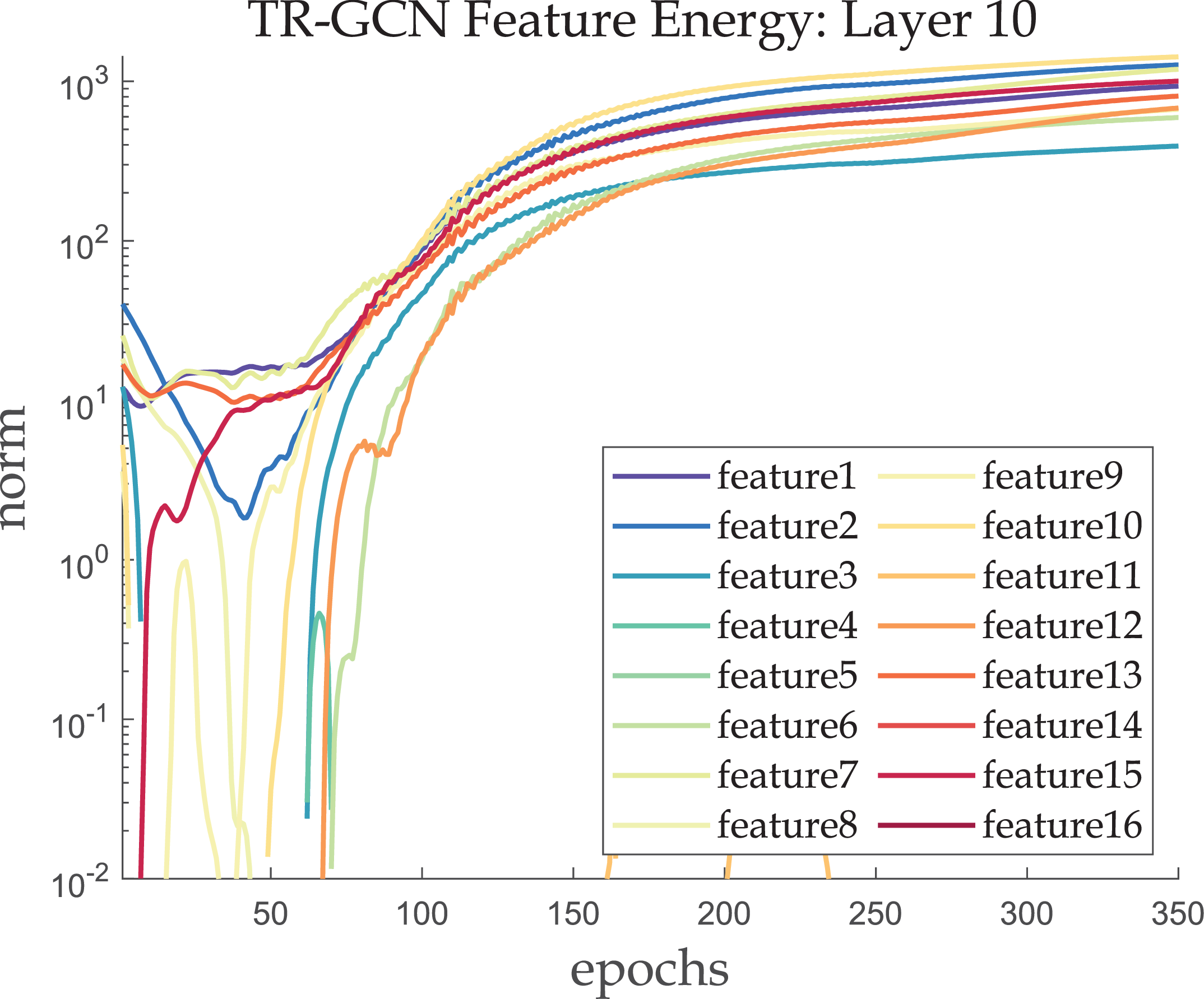}}
\caption{Comparison of energy changes in hidden layers of GCN and TR-GCN ($r=1$) during training}
\label{fig:Energy_comparison}
\end{figure*}

Through the forward analysis of energy flow in deep GCN, we can see that the energy of column features should reduce in top layers. But from Figure \ref{fig:Energy_comparison} (a)(b)(c) yielded by a numerical test with $10$-layer GCN, we can see that the energy of column features in top layers (Figure \ref{fig:Energy_comparison} (c)) do not have significant changes, while in bottom layers (Figure \ref{fig:Energy_comparison} (a)(b)) the energy of features shrinks during training. The cause of this contradiction is that we either have neglected \cite{li2018deeper} or have put too strong assumptions \cite{luan2019break,oono2019graph} on parameter matrices in forward analysis while ignore how parameter matrices changes in backpropagation. In the following, we will try to do gradient analysis from backward view and explain the energy loss in bottom layers in deep GCN.

\subsection{Backward Pass Analyses: Identifying the Core Problem}

\begin{figure*}[htbp]
\centering
\subfloat[]{
\captionsetup{justification = centering}
\includegraphics[width=0.24\textwidth]{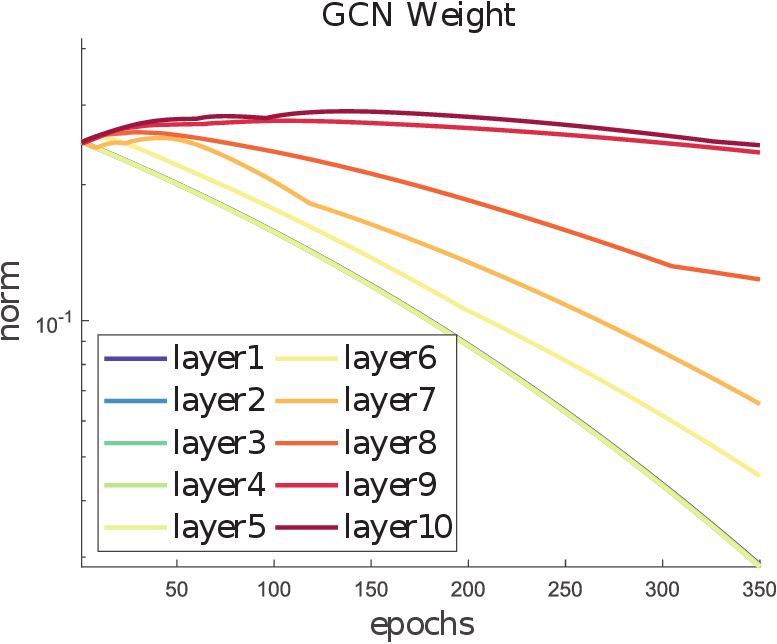}}
\hfill
\subfloat[]{
\captionsetup{justification = centering}
\includegraphics[width=0.24\textwidth]{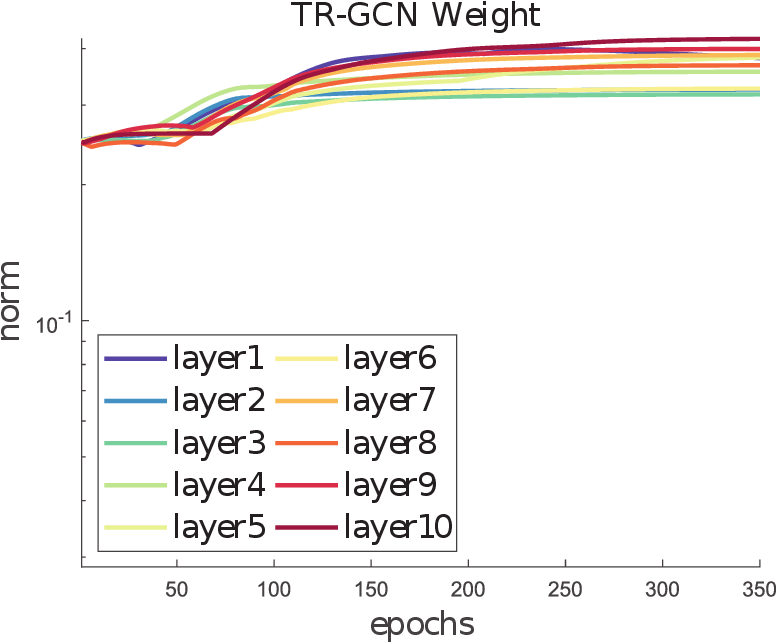}}
\hfill
\subfloat[]{
\captionsetup{justification = centering}
\includegraphics[width=0.24\textwidth]{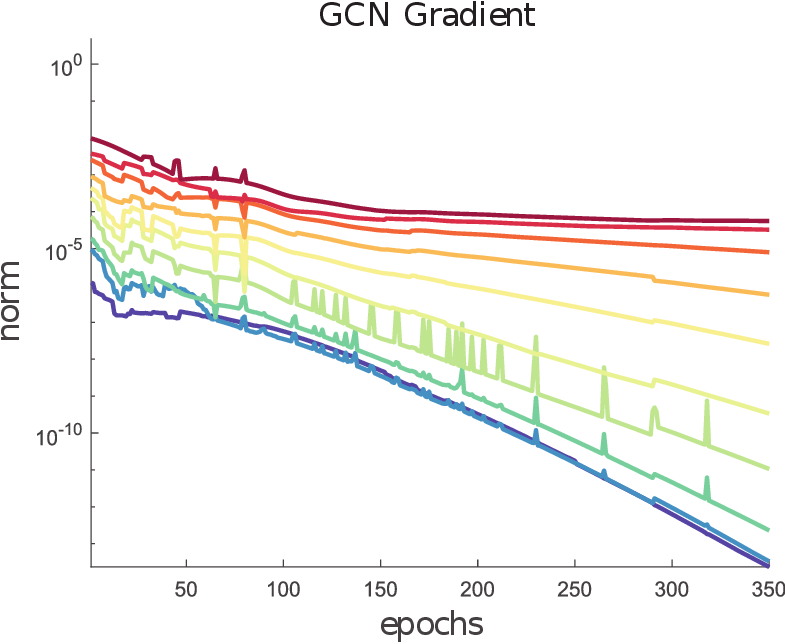}}
\hfill
\subfloat[]{
\captionsetup{justification = centering}
\includegraphics[width=0.24\textwidth]{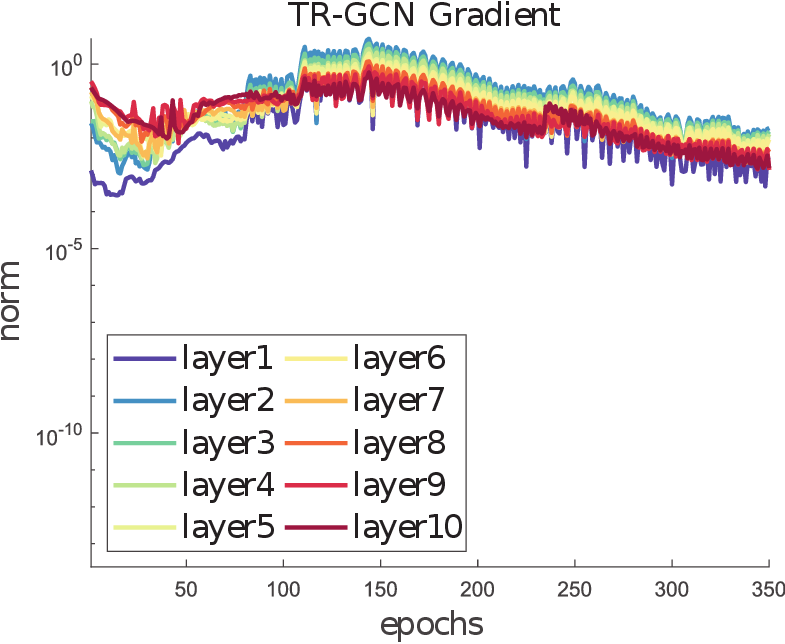}}
\caption{Comparison of weight and gradient norm in hidden layers of GCN and TR-GCN ($r=1$): the pairs have the same x- and y-ranges.}
\label{fig:weight_grad_comparison}
\end{figure*}

We first decompose the deep GCN architecture as follows
\begin{equation}\label{eq:deep_gcn_decompose}
\begin{aligned}
   & \bm{Y_0}  = \bm{X}, \ \ \bm{Y_1'} = \hat{A} \bm{X} W_0, \ \ \bm{Y_1} = \text{ReLU} (\hat{A} \bm{X} W_0) = \text{ReLU} (\bm{Y_1'}) = \bm{1}_{\mathbb{R}^+} (\bm{Y_1'}) \odot \bm{Y_1'}\\
   & \bm{Y_{i+1}'}  = \hat{A} \bm{Y_i} W_i, \ \ \bm{Y_{i+1}} = \text{ReLU} (\bm{Y_{i+1}'}) = \bm{1}_{\mathbb{R}^+} (\bm{Y_{i+1}'}) \odot \bm{Y_{i+1}'}, \ \ i = 1, \dots, n\\
   & \bm{Y} = \text{softmax} (\hat{A}  \bm{Y_n} W_n ) \equiv  \text{softmax} (\bm{Y'}) , \ \ l = -\text{trace} (\bm{Z}^T \text{log} \bm{Y})
\end{aligned}
\end{equation}

where $\bm{1}_{\mathbb{R}^+}(\cdot)$ and $\text{log}(\cdot)$ are pointwise indicator and log functions; $\odot$ is the Hadamard product; $\bm{Z}\in\Rbb^{N\times C}$ is the ground truth matrix with one-hot label vector $\bm{Z}_{i,:}$ in each row, $C$ is number of classes;  $l$ is the scalar loss. Then the gradient propagates in the following way,
\begin{equation}\label{eq:gradient}
\begin{aligned}
  \resizebox{1\hsize}{!}{ $\text{Output Layer } \ \frac{\partial l }{\partial \bm{Y'}} = \text{softmax}(\bm{Y'}) - \bm{Z}, \ \ \frac{\partial l }{\partial W_n} = \bm{Y_n}^T \hat{A} \frac{\partial l }{ \partial \bm{Y'}}, \ \ \frac{\partial l }{\partial \bm{Y_n}} = \hat{A} \frac{\partial l }{ \partial \bm{Y'}}  W_n^T$} \\
   \resizebox{1\hsize}{!}{ $\text{Hidden Layers } \ \frac{\partial l }{\partial \bm{Y_i'} } = \frac{\partial l }{\partial \bm{Y_i}} \odot \bm{1}_{\mathbb{R}^+}(\bm{Y_i'}), \ \ \frac{\partial l }{\partial W_{i-1}} = \bm{Y_{i-1}}^T \hat{A} \frac{\partial l }{\partial \bm{Y_i'} }, \ \ \frac{\partial l }{\partial \bm{Y_{i-1}}} = \hat{A} \frac{\partial l }{ \partial \bm{Y_i'}}  W_{i-1}^T$} 
\end{aligned}
\end{equation}
The gradient propagation of GCN differs from that of multi-layer perceptron (MLP)
by an extra multiplication of $\hat{A}$ when the gradient signal flows through $\bm{Y_i}$. Since $\abs{\lambda_i(\hat{A})} \leq 1$, this multiplication will cause energy loss of gradient signal (see Figure \ref{fig:weight_grad_comparison}(c)). In addition, oversmoothing does not only happen in feed-forward process, but also exists in backpropagation when we see $\frac{\partial l }{\partial \bm{Y_{i-1}}} = \hat{A} \frac{\partial l }{ \partial \bm{Y_i'}}  W_{i-1}^T$ as a backward view of hidden layers as \eqref{eq:deep_gcn_decompose}. In forward view, parameter matrix $W_i$ is fixed and we update $\Y_i$; in backward view, $\bm{Y_i}$ is fixed and we update $W_i$.  And the difference is in forward view, the input $\bm{X}$ is a fixed feature matrix, but in backward view, the scale of the input $\frac{\partial l }{\partial \bm{Y'}} = \text{softmax}(\bm{Y'}) - \bm{Z}$ (the prediction error) is getting smaller during training. Thus, the energy loss is more significant in $W_i$ (see Figure \ref{fig:weight_grad_comparison}(a)) from backward view and is more serious in bottom layers instead of in top layers.

This energy losing phenomenon is not an expressive power problem but a training issue. But this does not mean the training issue is the root cause of the performance limit problem, which we will draw conclusion later. In the following section, we propose method to alleviate the energy loss.

\section{Methods to Alleviate BP Energy Loss} \label{sec:alleviate_methods}

In this section, we propose $4$ methodologies to handle the problem of BP energy loss: \textit{spectra shift}, \textit{weight initialization}, \textit{normalization} and \textit{skip (residual) connection}.

\subsection{Spectra Shift and Topology Rescaling (TR)}

From the analysis in section \ref{sec:energy_gradient_analysis} and theorem 1, we can that $\abs{\lambda_i(\hat{A})} \leq 1$ is one of the main reasons of energy losing. To adjust $\abs{\lambda_i(\hat{A})}$ while maintaining certain topological properties of the original graph associated with  $\hat{A}$  (\eg{} the graph Fourier basis, the gap between the frequencies), we shift the spectra of $\hat{A}$ by changing $\hat{A}$ to $\hat{A}_r = r I +\hat{A}$, where $r$ is a real scalar.

\paragraph{Physical Meaning} Spectra shift is also a commonly used method in community detection  \cite{arenas2008analysis} to address the so-called ``resolution limit'' challenge \cite{fortunato2007resolution,good2010performance,zhang2009modularity}, \ie{} it can only produce the modules at a certain characteristic scale \cite{xiang2015multi} while is unable to extract densely connected substructures with small sizes. Spectra shift rescales the graph topology with a proper self-loop assignment through which we can adjust the strength (degree) of each node \cite{xiang2015multi} and $r$ is named resolution parameter. The translation of strengths has no impact on the original connection of nodes, which are the building blocks of the topology. The shift only balance the the property of each node individually and in the same way for all of them. 

Spectra shift essentially allows the graph operator to adjusts the scale of the components of graph signal in graph frequency domain. To see this, suppose that $\hat{A}$ has the eigendecomposition 
$\hat{A} = \hat{U} \hat{\Lambda} \hat{U}^T$, where $\hat{U}$ is orthogonal.
Then
\begin{equation} 
    \x =  \sum_i  \hat{\bm{u}}_i (\hat{\bm{u}}_i ^T \bm{x}), \ \ 
    \hat{A} \bm{x} = \sum_i \hat{\lambda}_i  \hat{\bm{u}}_i (\hat{\bm{u}}_i ^T \bm{x}), \ \ 
    \hat{A}_r \bm{x} = \sum_i (\hat{\lambda}_i+r)  \hat{\bm{u}}_i (\hat{\bm{u}}_i ^T \bm{x})
    \label{eq:explicit_conv_L}
\end{equation}
Note that the components of $\x$, $\hat{A}\x$ and $\hat{A}_r\x$ in the direction $\hat{\u}_i$ 
are $\hat{\bm{u}}_i^T \bm{x}$,  $\hat{\lambda}_i\hat{\bm{u}}_i^T \bm{x}$,
and  $(\hat{\lambda}_i+r)\tilde{\bm{u}}_i^T \bm{x}$, respectively.
Thus applying the operator $\hat{A}$ to $\x$ just scales the component of $\x$ in the direction of $\u_i$
by $\hat{\lambda}_i$ for each $i$.

Tuning the resolution parameter $r$ actually rescales those components in the way that global information (high smoothness) will be increased with positive $r$, and local information (low smoothness) will be enhanced with negative $r$. The GCN with $\hat{A}_r$ is called topology rescaling GCN (TR-GCN).

Note that $\lambda_i(\hat{A}) \in (-1,1]$.
A shift which makes  ${\max}_i \abs{\lambda_i(\hat{A}) +r} \geq 1$
is considered risky because it will cause gradient exploding and numerical instability during training as stated in \cite{kipf2016classification}. However, through our analysis, TR-GCN will not only overcome the difficulty when training in deep architecture (see figure \ref{fig:Energy_comparison}(d)(e)(f) and figure \ref{fig:weight_grad_comparison}(b)(d)), but also will not lose expressive power (see table \ref{tab:ablation}) by setting a proper $r$ (depends on the task and size of the network).

\subsection{Weight Initialization}

The gradient propagation does not only depends on $\hat{A}$ but also depends on the scale of $W_i$. An initialization with proper scale would make $W_i$ get undiminished gradient from the start of training and move to the correct direction with a clearer signal \cite{luan2023addressing}. Thus, we adjust the scale of each element in $W_i$ initialized by \cite{glorot2010understanding} with a tunable constant $\lambda_{\text{init}}$ as follows,
\begin{equation}
 \lambda_{\text{init}} \times U(-\frac{1}{\sqrt{F_{i+1}}}, \frac{1}{\sqrt{F_{i+1}}}) \; \text{ or } \; \lambda_{\text{init}} \times N(0, \sqrt{\frac{2}{F_i + F_{i+1}}} )
\end{equation}

\subsection{Normalization}

Normalization is a natural method to control the energy flow. A direct method would be to normalize the output matrix in each hidden layer with a constant $\lambda_E$; Or, an indirect method can also be used to normalize the weight matrix \cite{salimans2016weight} by a constant $\lambda_W$, which shares the same spirit of normalizing the largest singular value of $W_i$ \cite{oono2019graph}.
\begin{equation}
\begin{aligned}
\text{Energy Normalization: }& \bm{Y_i} = \lambda_{E} \cdot  \bm{Y_i}/\norm{\bm{Y_i}}_2\\
\text{Weight Normalization: }& W_i = \lambda_{W} \cdot W_i/\norm{W_i}_2\\
\end{aligned}
\end{equation}

\subsection{Skip Connection}

Skip (residual) connections \cite{he2016deep} is a widely used technique in training deep neural networks and has achieved success in feature extraction. It helps with gradient propagation without introducing additional parameters. Skip connections, if adapted in GCNs, will have the general form as follows:

\begin{equation}
\begin{aligned}
\bm{Y}_{i+1} = \bm{Y}_i + \sigma(\hat{A}\bm{Y}_i W_i)
\end{aligned}
\end{equation}
where $\sigma$ is the activation function, $\bm{Y}_i$ is the input of the $i$-th layer and $\bm{Y}_{i+1}$ is the output of the $i$-th layer as well as the input of the $(i+1)$-th layer.

It is shown that existing GCN models are difficult to train when they are scaled with more than $7$-layer-deep. This is possible due to the increase of the effective context size of each node and overfitting issue as stated in \cite{kipf2016classification}. There exists one method ResGCN \cite{li2019can} that seeks also to address such problem via residual connections, but it actually uses concatenation of the intermediate outputs of hidden layers, introducing excessive parameters. The effectiveness shown in experiments are actually not only the result of the skip-connections but also the expressive power of additional parameters. However, in experiments, we will show that residual connections alone could accomplish the task.

\section{Experiments}\label{sec:experiments}

This section is crucial to the paper's main hypothesis: can we boost the performance of GNNs by just training them better? For this purpose, we patch the most-popular baseline GCN with the ideas in previous section to form a set of detailed comparative tests and fix the architecture to be $10$-layers deep throughout the entire section\footnote{The source code will be submitted within the supplementary materials for blind review and open-source afterwards.}. Particularly, we have selected the node classification tasks on Cora, CiteSeer and PubMed, the three most popular datasets. We use the most classic setting on training, which is identical to the one suggested in \cite{yang2016revisiting}. The section features two sets of experiments, the first of which validates the effectiveness of the proposed methods lowering the training difficulty while the second demonstrates the potential performance boost when the patched methods are fine-tuned. For all experiments, we used Adam optimizer and ReLU as the activation function (PyTorch implementation).

\subsection{Training Difficulty \& Generalization}

Instead of demonstrating how good the performance of the patched method could possibly be, the first set of experiments focuses on validating the effectiveness of the proposed ideas aiming to lower the training difficulty with a detailed ablation study. Also, we investigate the potential loss of generalization abilities, \ie{} whether these ideas lead to overfitting.

For fair comparison, we use the same base architecture for the baseline and all the patched methods: 10 GCN layers each with width 16. Also, we utilize the same set of basic hyperparameters: a learning rate of 0.001, weight decay of $5\times 10^{-4}$, $0$ dropout. We train all methods to the same extent by using the same training procedures for all the methods: each method in each run is trained until the validation loss is not improved for $200$ epochs.

With these, we run each method on Cora dataset with public split (20 training data for each class) for $20$ independent runs and obtain the final reported classification accuracy together with the standard deviation of the accuracy. The results also include the errors (losses) computed on the training set and the test set. The results are reported in Table \ref{tab:ablation}, together with the hyperparameters included additionally by the patched methods. Note that these hyperparameters are not fine-tuned. Also, since all methods are trained with the same base loss (negative log-likelihood) and additional losses introduced by the patched methods only increase the total loss, the comparison of loss among the methods can fairly tell that the patched methods' ability of lowering the training difficulty if their training losses are lower than the baseline.

\begin{table}[htbp]
\setlength{\tabcolsep}{1pt}
  \centering
  \caption{ Ablation Tests for Training Difficulties on Cora}
  \resizebox{1\hsize}{!}{
    \begin{tabular}{cccccccccccccc}
    \toprule
    \toprule
    \multicolumn{2}{c}{\textit{Train Loss}} & \multicolumn{2}{c}{\textit{Train Acc}} & \multicolumn{2}{c}{\textit{Test Loss}} & \multicolumn{2}{c}{\textit{Test Acc}} & \cellcolor[rgb]{ .608,  .761,  .902}Change L & \cellcolor[rgb]{ .663,  .51,  .871}Change All & \multicolumn{4}{c}{\cellcolor[rgb]{ .957,  .69,  .518}Change W, b} \\
    Mean  & Std   & Mean  & Std   & Mean  & Std   & Mean  & Std   & resolution & skip  & weight norm & energy norm & weight init & weight const \\
    \midrule
    \rowcolor[rgb]{ .973,  .412,  .42} 1.946 & \cellcolor[rgb]{ 1,  1,  1}0.000 & 14.29\% & \cellcolor[rgb]{ 1,  1,  1}0.00\% & \cellcolor[rgb]{ 1,  .871,  .51}1.960 & \cellcolor[rgb]{ 1,  1,  1}0.035 & 23.11\% & \cellcolor[rgb]{ 1,  1,  1}8.80\% & \cellcolor[rgb]{ 1,  1,  1}~ & \cellcolor[rgb]{ 1,  1,  1}N & \cellcolor[rgb]{ 1,  1,  1}~ & \cellcolor[rgb]{ 1,  1,  1}~ & \cellcolor[rgb]{ 1,  1,  1}uniform & \cellcolor[rgb]{ 1,  1,  1}~ \\
    \rowcolor[rgb]{ .459,  .765,  .486} \textit{0.004} & \cellcolor[rgb]{ 1,  1,  1}\textit{0.008} & \cellcolor[rgb]{ 1,  .922,  .518}\textit{99.93\%} & \cellcolor[rgb]{ 1,  1,  1}\textit{0.21\%} & \cellcolor[rgb]{ .973,  .412,  .42}\textit{4.608} & \cellcolor[rgb]{ 1,  1,  1}\textit{3.244} & \cellcolor[rgb]{ .992,  .796,  .49}\textit{57.10\%} & \cellcolor[rgb]{ 1,  1,  1}\textit{7.85\%} & \cellcolor[rgb]{ .867,  .922,  .969}1.00 & \cellcolor[rgb]{ 1,  1,  1}N & \cellcolor[rgb]{ 1,  1,  1}~ & \cellcolor[rgb]{ 1,  1,  1}~ & \cellcolor[rgb]{ 1,  1,  1}uniform & \cellcolor[rgb]{ 1,  1,  1}~ \\
    \rowcolor[rgb]{ 1,  .902,  .514} \textit{0.106} & \cellcolor[rgb]{ 1,  1,  1}\textit{0.101} & \cellcolor[rgb]{ .996,  .91,  .514}\textit{98.07\%} & \cellcolor[rgb]{ 1,  1,  1}\textit{2.14\%} & \cellcolor[rgb]{ 1,  .894,  .514}\textit{1.806} & \cellcolor[rgb]{ 1,  1,  1}\textit{0.485} & \cellcolor[rgb]{ .996,  .914,  .514}\textit{67.45\%} & \cellcolor[rgb]{ 1,  1,  1}\textit{4.37\%} & \cellcolor[rgb]{ 1,  1,  1}~ & \cellcolor[rgb]{ 1,  1,  1}N & \cellcolor[rgb]{ 1,  1,  1}~ & \cellcolor[rgb]{ 1,  1,  1}~ & \cellcolor[rgb]{ .957,  .69,  .518}normal & \cellcolor[rgb]{ .957,  .69,  .518}1.8 \\
    \rowcolor[rgb]{ .502,  .776,  .486} \textit{0.005} & \cellcolor[rgb]{ 1,  1,  1}\textit{0.010} & \cellcolor[rgb]{ .388,  .745,  .482}\textit{100.00\%} & \cellcolor[rgb]{ 1,  1,  1}\textit{0.00\%} & \cellcolor[rgb]{ .992,  .706,  .478}\textit{2.908} & \cellcolor[rgb]{ 1,  1,  1}\textit{1.838} & \cellcolor[rgb]{ .996,  .89,  .51}\textit{65.13\%} & \cellcolor[rgb]{ 1,  1,  1}\textit{4.53\%} & \cellcolor[rgb]{ .867,  .922,  .969}1.00 & \cellcolor[rgb]{ 1,  1,  1}N & \cellcolor[rgb]{ 1,  1,  1}~ & \cellcolor[rgb]{ 1,  1,  1}~ & \cellcolor[rgb]{ .957,  .69,  .518}normal & \cellcolor[rgb]{ .957,  .69,  .518}0.8 \\
    \rowcolor[rgb]{ .992,  .714,  .478} \textit{0.811} & \cellcolor[rgb]{ 1,  1,  1}\textit{0.795} & \cellcolor[rgb]{ .988,  .753,  .482}\textit{71.93\%} & \cellcolor[rgb]{ 1,  1,  1}\textit{28.03\%} & \cellcolor[rgb]{ .557,  .792,  .49}\textit{1.184} & \cellcolor[rgb]{ 1,  1,  1}\textit{0.306} & \cellcolor[rgb]{ .996,  .882,  .51}\textit{64.68\%} & \cellcolor[rgb]{ 1,  1,  1}\textit{9.94\%} & \cellcolor[rgb]{ 1,  1,  1}~ & \cellcolor[rgb]{ 1,  1,  1}N & \cellcolor[rgb]{ .957,  .69,  .518}7 & \cellcolor[rgb]{ 1,  1,  1}~ & \cellcolor[rgb]{ 1,  1,  1}uniform & \cellcolor[rgb]{ 1,  1,  1}~ \\
    \rowcolor[rgb]{ .675,  .827,  .498} \textit{0.011} & \cellcolor[rgb]{ 1,  1,  1}\textit{0.011} & \cellcolor[rgb]{ .388,  .745,  .482}\textit{100.00\%} & \cellcolor[rgb]{ 1,  1,  1}\textit{0.00\%} & \cellcolor[rgb]{ .463,  .765,  .486}\textit{1.088} & \cellcolor[rgb]{ 1,  1,  1}\textit{0.105} & \cellcolor[rgb]{ .816,  .871,  .51}\textit{69.72\%} & \cellcolor[rgb]{ 1,  1,  1}\textit{2.55\%} & \cellcolor[rgb]{ 1,  1,  1}~ & \cellcolor[rgb]{ 1,  1,  1}N & \cellcolor[rgb]{ 1,  1,  1}~ & \cellcolor[rgb]{ .957,  .69,  .518}800 & \cellcolor[rgb]{ 1,  1,  1}uniform & \cellcolor[rgb]{ 1,  1,  1}~ \\
    \rowcolor[rgb]{ .996,  .835,  .502} \textit{0.359} & \cellcolor[rgb]{ 1,  1,  1}\textit{0.361} & \cellcolor[rgb]{ .996,  .859,  .506}\textit{89.79\%} & \cellcolor[rgb]{ 1,  1,  1}\textit{16.61\%} & \cellcolor[rgb]{ .922,  .898,  .51}\textit{1.562} & \cellcolor[rgb]{ 1,  1,  1}\textit{0.328} & \cellcolor[rgb]{ .996,  .878,  .51}\textit{64.35\%} & \cellcolor[rgb]{ 1,  1,  1}\textit{4.65\%} & \cellcolor[rgb]{ .867,  .922,  .969}1.00 & \cellcolor[rgb]{ 1,  1,  1}N & \cellcolor[rgb]{ .957,  .69,  .518}5 & \cellcolor[rgb]{ 1,  1,  1}~ & \cellcolor[rgb]{ 1,  1,  1}uniform & \cellcolor[rgb]{ 1,  1,  1}~ \\
    \rowcolor[rgb]{ .404,  .749,  .482} \textit{0.002} & \cellcolor[rgb]{ 1,  1,  1}\textit{0.003} & \cellcolor[rgb]{ .388,  .745,  .482}\textit{100.00\%} & \cellcolor[rgb]{ 1,  1,  1}\textit{0.00\%} & \cellcolor[rgb]{ 1,  .878,  .51}\textit{1.912} & \cellcolor[rgb]{ 1,  1,  1}\textit{0.627} & \cellcolor[rgb]{ .996,  .855,  .502}\textit{62.08\%} & \cellcolor[rgb]{ 1,  1,  1}\textit{2.95\%} & \cellcolor[rgb]{ .867,  .922,  .969}1.00 & \cellcolor[rgb]{ 1,  1,  1}N & \cellcolor[rgb]{ 1,  1,  1}~ & \cellcolor[rgb]{ .957,  .69,  .518}550 & \cellcolor[rgb]{ 1,  1,  1}uniform & \cellcolor[rgb]{ 1,  1,  1}~ \\
    \rowcolor[rgb]{ .584,  .8,  .49} \textit{0.008} & \cellcolor[rgb]{ 1,  1,  1}\textit{0.009} & \cellcolor[rgb]{ 1,  .922,  .518}\textit{99.93\%} & \cellcolor[rgb]{ 1,  1,  1}\textit{0.21\%} & \cellcolor[rgb]{ 1,  .91,  .518}\textit{1.723} & \cellcolor[rgb]{ 1,  1,  1}\textit{1.045} & \cellcolor[rgb]{ .973,  .914,  .518}\textit{68.15\%} & \cellcolor[rgb]{ 1,  1,  1}\textit{5.29\%} & \cellcolor[rgb]{ .867,  .922,  .969}1.00 & \cellcolor[rgb]{ .663,  .51,  .871}Y & \cellcolor[rgb]{ 1,  1,  1}~ & \cellcolor[rgb]{ 1,  1,  1}~ & \cellcolor[rgb]{ 1,  1,  1}uniform & \cellcolor[rgb]{ 1,  1,  1}~ \\
    \rowcolor[rgb]{ 1,  .922,  .518} \textit{0.034} & \cellcolor[rgb]{ 1,  1,  1}\textit{0.024} & \cellcolor[rgb]{ .996,  .918,  .514}\textit{99.79\%} & \cellcolor[rgb]{ 1,  1,  1}\textit{0.46\%} & \cellcolor[rgb]{ .686,  .827,  .498}\textit{1.318} & \cellcolor[rgb]{ 1,  1,  1}\textit{0.530} & \cellcolor[rgb]{ .557,  .796,  .494}\textit{72.30\%} & \cellcolor[rgb]{ 1,  1,  1}\textit{2.59\%} & \cellcolor[rgb]{ 1,  1,  1}~ & \cellcolor[rgb]{ .663,  .51,  .871}Y & \cellcolor[rgb]{ 1,  1,  1}~ & \cellcolor[rgb]{ 1,  1,  1}~ & \cellcolor[rgb]{ .957,  .69,  .518}normal & \cellcolor[rgb]{ .957,  .69,  .518}0.9 \\
    \rowcolor[rgb]{ .996,  .827,  .502} \textit{0.378} & \cellcolor[rgb]{ 1,  1,  1}\textit{0.194} & \cellcolor[rgb]{ .996,  .894,  .51}\textit{95.43\%} & \cellcolor[rgb]{ 1,  1,  1}\textit{2.77\%} & \cellcolor[rgb]{ .388,  .745,  .482}\textit{1.009} & \cellcolor[rgb]{ 1,  1,  1}\textit{0.146} & \cellcolor[rgb]{ .388,  .745,  .482}\textit{73.98\%} & \cellcolor[rgb]{ 1,  1,  1}\textit{2.68\%} & \cellcolor[rgb]{ 1,  1,  1}~ & \cellcolor[rgb]{ .663,  .51,  .871}Y & \cellcolor[rgb]{ .957,  .69,  .518}7 & \cellcolor[rgb]{ 1,  1,  1}~ & \cellcolor[rgb]{ 1,  1,  1}uniform & \cellcolor[rgb]{ 1,  1,  1}~ \\
    \rowcolor[rgb]{ .439,  .757,  .482} \textit{0.003} & \cellcolor[rgb]{ 1,  1,  1}\textit{0.003} & \cellcolor[rgb]{ .388,  .745,  .482}\textit{100.00\%} & \cellcolor[rgb]{ 1,  1,  1}\textit{0.00\%} & \cellcolor[rgb]{ .902,  .89,  .51}\textit{1.543} & \cellcolor[rgb]{ 1,  1,  1}\textit{0.488} & \cellcolor[rgb]{ .659,  .824,  .498}\textit{71.28\%} & \cellcolor[rgb]{ 1,  1,  1}\textit{2.95\%} & \cellcolor[rgb]{ 1,  1,  1}~ & \cellcolor[rgb]{ .663,  .51,  .871}Y & \cellcolor[rgb]{ 1,  1,  1}~ & \cellcolor[rgb]{ .957,  .69,  .518}2900 & \cellcolor[rgb]{ 1,  1,  1}uniform & \cellcolor[rgb]{ 1,  1,  1}~ \\
    \rowcolor[rgb]{ .388,  .745,  .482} \textit{0.001} & \cellcolor[rgb]{ 1,  1,  1}\textit{0.001} & \textit{100.00\%} & \cellcolor[rgb]{ 1,  1,  1}\textit{0.00\%} & \cellcolor[rgb]{ 1,  .867,  .51}\textit{1.969} & \cellcolor[rgb]{ 1,  1,  1}\textit{0.811} & \cellcolor[rgb]{ .996,  .918,  .514}\textit{67.58\%} & \cellcolor[rgb]{ 1,  1,  1}\textit{4.93\%} & \cellcolor[rgb]{ .867,  .922,  .969}1.00 & \cellcolor[rgb]{ .663,  .51,  .871}Y & \cellcolor[rgb]{ 1,  1,  1}~ & \cellcolor[rgb]{ 1,  1,  1}~ & \cellcolor[rgb]{ 1,  1,  1}uniform & \cellcolor[rgb]{ 1,  1,  1}~ \\
    \rowcolor[rgb]{ .514,  .78,  .486} \textit{0.005} & \cellcolor[rgb]{ 1,  1,  1}\textit{0.003} & \cellcolor[rgb]{ .388,  .745,  .482}\textit{100.00\%} & \cellcolor[rgb]{ 1,  1,  1}\textit{0.00\%} & \cellcolor[rgb]{ .808,  .867,  .506}\textit{1.447} & \cellcolor[rgb]{ 1,  1,  1}\textit{0.498} & \cellcolor[rgb]{ .855,  .882,  .51}\textit{69.32\%} & \cellcolor[rgb]{ 1,  1,  1}\textit{2.83\%} & \cellcolor[rgb]{ .867,  .922,  .969}1.00 & \cellcolor[rgb]{ .663,  .51,  .871}Y & \cellcolor[rgb]{ 1,  1,  1}~ & \cellcolor[rgb]{ 1,  1,  1}~ & \cellcolor[rgb]{ .957,  .69,  .518}normal & \cellcolor[rgb]{ .957,  .69,  .518}0.5 \\
    \rowcolor[rgb]{ 1,  .878,  .51} \textit{0.193} & \cellcolor[rgb]{ 1,  1,  1}\textit{0.115} & \cellcolor[rgb]{ .996,  .91,  .514}\textit{98.50\%} & \cellcolor[rgb]{ 1,  1,  1}\textit{1.08\%} & \cellcolor[rgb]{ .616,  .808,  .494}\textit{1.247} & \cellcolor[rgb]{ 1,  1,  1}\textit{0.233} & \cellcolor[rgb]{ .941,  .906,  .518}\textit{68.48\%} & \cellcolor[rgb]{ 1,  1,  1}\textit{3.72\%} & \cellcolor[rgb]{ .867,  .922,  .969}1.00 & \cellcolor[rgb]{ .663,  .51,  .871}Y & \cellcolor[rgb]{ .957,  .69,  .518}3 & \cellcolor[rgb]{ 1,  1,  1}~ & \cellcolor[rgb]{ 1,  1,  1}uniform & \cellcolor[rgb]{ 1,  1,  1}~ \\
    \rowcolor[rgb]{ 1,  .91,  .518} \textit{0.080} & \cellcolor[rgb]{ 1,  1,  1}\textit{0.041} & \cellcolor[rgb]{ .388,  .745,  .482}\textit{100.00\%} & \cellcolor[rgb]{ 1,  1,  1}\textit{0.00\%} & \cellcolor[rgb]{ .996,  .851,  .506}\textit{2.074} & \cellcolor[rgb]{ 1,  1,  1}\textit{0.218} & \cellcolor[rgb]{ .737,  .847,  .506}\textit{70.52\%} & \cellcolor[rgb]{ 1,  1,  1}\textit{2.39\%} & \cellcolor[rgb]{ .867,  .922,  .969}1.00 & \cellcolor[rgb]{ .663,  .51,  .871}Y & \cellcolor[rgb]{ 1,  1,  1}~ & \cellcolor[rgb]{ .957,  .69,  .518}325 & \cellcolor[rgb]{ 1,  1,  1}uniform & \cellcolor[rgb]{ 1,  1,  1}~ \\
    \bottomrule
    \bottomrule
    \multicolumn{14}{l}{\tiny Each row represents a method. The first four columns are featured with color indicators: the greener the better result, the redder the worse. The changes applied unto the baseline are highlighted in the later columns. The first row has no colored changes and is therefore the baseline.}\\
    \multicolumn{14}{l}{\tiny We use different highlight colors to indicate the change on the operators: blue for the changes on graph operator L, red for the changes on $W$ and $\bm{b}$ and purple (blue + red) for the changes applied on all $L$, $W$ and $\bm{b}$.}
    \end{tabular}%
    }
  \label{tab:ablation}%
\end{table}%

From the results on the training set, we can observe significantly smaller training loss (more than 50\%) and significantly higher training accuracy (more than 6 times), comparing those of the patched methods and the original baseline. Considering that all of the compared methods have exactly the same parameter composition, we can safely say that the proposed methods are indeed effective lowering the training difficulties. However, we cannot conclude from the results which single idea contributes the most to the training difficulty alleviation. 

Comparing the results on the test set, we can see that the error and accuracy on the test set ruled out the argument of overfitting: generally all the losses and accuracy on the test set are improved significantly. With all the observations in this set of experiments, the validation of the hypothesis is finished: we can make GNNs perform better by training them better.

\subsection{Finetuned Performance Boost}

In this second set of experiments, we fine tune each method (including the baseline) and compare their best reported performance. This shows how much potential could be unlocked by better training procedures. The fine-tuning is conducted with Bayesian optimization \cite{shahriari2016bayesian} to the same extent\footnote{All methods are fixed $10$-layer deep. Methods share the same search range for the base hyperparameters (learning rate in $[10^{-6}, 10^{-1}]$, weight decay in $[10^{-5}, 10^{-1}]$, width in $\{100, 200, \dots, 5000\}$, dropout in $(0, 1)$). The hyperparameters unique to the patched methods are also fixed for each patched method (resolution in $[-1, 5]$, weight constant in $[0.1, 5]$, weight normalization coefficient in $[1, 15]$, energy normalization coefficient in $[25, 2500]$). The search stops if the performance is not improved for $64$ candidates.}. Each result reported in Table \ref{tab:finetune} is averaged from $20$ independent runs together with the standard deviation\footnote{GCN is reproduced and performed fine-tuning upon.}.

\begin{table}[htbp]
\setlength{\tabcolsep}{4pt}
  \centering
  \caption{Fine-tuned Performance on Node Classification Tasks}
  \resizebox{1\hsize}{!}{
    \begin{tabular}{ccccccccccc}
    \toprule
    \toprule
    \multicolumn{2}{c}{\textit{Cora}} & \multicolumn{2}{c}{\textit{CiteSeer}} & \multicolumn{2}{c}{\textit{PubMed}} & \cellcolor[rgb]{ .608,  .761,  .902}Change L & \cellcolor[rgb]{ .663,  .51,  .871}Change All & \multicolumn{3}{c}{\cellcolor[rgb]{ .957,  .69,  .518}Change W, b} \\
    Mean  & Std   & Mean  & Std   & Mean  & Std   & resolution & skip  & weight norm & energy norm & weight init \\
    \midrule
    \rowcolor[rgb]{ .973,  .412,  .42} 74.66\% & \cellcolor[rgb]{ 1,  1,  1}1.37\% & 60.39\% & \cellcolor[rgb]{ 1,  1,  1}2.67\% & 74.01\% & \cellcolor[rgb]{ 1,  1,  1}1.40\% & \cellcolor[rgb]{ 1,  1,  1}~ & \cellcolor[rgb]{ 1,  1,  1}~ & \cellcolor[rgb]{ 1,  1,  1}~ & \cellcolor[rgb]{ 1,  1,  1}~ & \cellcolor[rgb]{ 1,  1,  1}uniform \\
    \rowcolor[rgb]{ .996,  .867,  .506} \textit{82.06\%} & \cellcolor[rgb]{ 1,  1,  1}\textit{0.56\%} & \cellcolor[rgb]{ .996,  .855,  .502}\textit{71.54\%} & \cellcolor[rgb]{ 1,  1,  1}\textit{2.54\%} & \cellcolor[rgb]{ .996,  .882,  .51}\textit{78.48\%} & \cellcolor[rgb]{ 1,  1,  1}\textit{1.52\%} & \cellcolor[rgb]{ .867,  .922,  .969} & \cellcolor[rgb]{ 1,  1,  1}~ & \cellcolor[rgb]{ 1,  1,  1}~ & \cellcolor[rgb]{ 1,  1,  1}~ & \cellcolor[rgb]{ 1,  1,  1}uniform \\
    \rowcolor[rgb]{ .388,  .745,  .482} \textit{\textbf{83.52\%}} & \cellcolor[rgb]{ 1,  1,  1}\textit{\textbf{0.91\%}} & \cellcolor[rgb]{ .565,  .796,  .494}\textit{73.64\%} & \cellcolor[rgb]{ 1,  1,  1}\textit{0.75\%} & \textit{\textbf{79.20\%}} & \cellcolor[rgb]{ 1,  1,  1}\textit{\textbf{1.16\%}} & \cellcolor[rgb]{ 1,  1,  1} & \cellcolor[rgb]{ .663,  .51,  .871} & \cellcolor[rgb]{ .957,  .69,  .518} & \cellcolor[rgb]{ 1,  1,  1}~ & \cellcolor[rgb]{ 1,  1,  1}uniform \\
    \rowcolor[rgb]{ .831,  .875,  .51} \textit{83.10\%} & \cellcolor[rgb]{ 1,  1,  1}\textit{0.84\%} & \cellcolor[rgb]{ .478,  .773,  .49}\textit{73.74\%} & \cellcolor[rgb]{ 1,  1,  1}\textit{1.00\%} & \cellcolor[rgb]{ .878,  .886,  .514}\textit{78.92\%} & \cellcolor[rgb]{ 1,  1,  1}\textit{0.77\%} & \cellcolor[rgb]{ 1,  1,  1} & \cellcolor[rgb]{ .663,  .51,  .871} & \cellcolor[rgb]{ 1,  1,  1}~ & \cellcolor[rgb]{ .957,  .69,  .518} & \cellcolor[rgb]{ 1,  1,  1}uniform \\
    \rowcolor[rgb]{ .98,  .918,  .518} \textit{82.96\%} & \cellcolor[rgb]{ 1,  1,  1}\textit{1.21\%} & \cellcolor[rgb]{ .388,  .745,  .482}\textit{\textbf{73.84\%}} & \cellcolor[rgb]{ 1,  1,  1}\textit{\textbf{0.82\%}} & \cellcolor[rgb]{ .996,  .91,  .514}\textit{78.76\%} & \cellcolor[rgb]{ 1,  1,  1}\textit{0.91\%} & \cellcolor[rgb]{ 1,  1,  1} & \cellcolor[rgb]{ .663,  .51,  .871} & \cellcolor[rgb]{ 1,  1,  1}~ & \cellcolor[rgb]{ 1,  1,  1}~ & \cellcolor[rgb]{ .957,  .69,  .518} \\
    \rowcolor[rgb]{ .388,  .745,  .482} \textit{83.52\%} & \cellcolor[rgb]{ 1,  1,  1}\textit{0.51\%} & \cellcolor[rgb]{ .863,  .882,  .51}\textit{73.30\%} & \cellcolor[rgb]{ 1,  1,  1}\textit{1.33\%} & \cellcolor[rgb]{ .741,  .847,  .506}\textit{79.00\%} & \cellcolor[rgb]{ 1,  1,  1}\textit{0.67\%} & \cellcolor[rgb]{ .867,  .922,  .969} & \cellcolor[rgb]{ .663,  .51,  .871} & \cellcolor[rgb]{ .957,  .69,  .518} & \cellcolor[rgb]{ 1,  1,  1}~ & \cellcolor[rgb]{ 1,  1,  1}uniform \\
    \rowcolor[rgb]{ .996,  .918,  .514} \textit{82.92\%} & \cellcolor[rgb]{ 1,  1,  1}\textit{0.71\%} & \cellcolor[rgb]{ .996,  .914,  .514}\textit{72.98\%} & \cellcolor[rgb]{ 1,  1,  1}\textit{1.34\%} & \cellcolor[rgb]{ .996,  .914,  .514}\textit{78.78\%} & \cellcolor[rgb]{ 1,  1,  1}\textit{0.59\%} & \cellcolor[rgb]{ .867,  .922,  .969} & \cellcolor[rgb]{ .663,  .51,  .871} & \cellcolor[rgb]{ 1,  1,  1}~ & \cellcolor[rgb]{ .957,  .69,  .518} & \cellcolor[rgb]{ 1,  1,  1}uniform \\
    \rowcolor[rgb]{ .996,  .871,  .506} \textit{82.16\%} & \cellcolor[rgb]{ 1,  1,  1}\textit{0.88\%} & \cellcolor[rgb]{ .996,  .851,  .502}\textit{71.40\%} & \cellcolor[rgb]{ 1,  1,  1}\textit{1.35\%} & \cellcolor[rgb]{ .741,  .847,  .506}\textit{79.00\%} & \cellcolor[rgb]{ 1,  1,  1}\textit{1.05\%} & \cellcolor[rgb]{ .867,  .922,  .969} & \cellcolor[rgb]{ .663,  .51,  .871} & \cellcolor[rgb]{ 1,  1,  1}~ & \cellcolor[rgb]{ 1,  1,  1}~ & \cellcolor[rgb]{ .957,  .69,  .518} \\
    \bottomrule
    \bottomrule
    \multicolumn{11}{l}{\tiny All the architectures are fixed with depth $10$.}
    \end{tabular}%
    }
  \label{tab:finetune}%
\end{table}%

From the results in the table, we observe that the patched methods obtain statistically significant performance boost. Therefore, together with the observations from the previous set of experiments, we conclude that the proposed methods could indeed alleviate the performance limit problem by lowering the training difficulty.

\section{Conclusion}\label{sec:conclusion}

In this paper, we verify the hypothesis that the cause of the performance limit problem of deep GCNs are more likely the training difficulty rather than insufficient capabilities. Out of the analyses on signal energy, we address the problem by proposing several methodologies that seek to mitigate the training process. The contribution enables lightweight GCN architectures to gain better performance when stacked deeper.

Though the proposed methods show effectiveness in lowering the training loss and improving the performance in practice, the methods introduce additional hyperparameters that require tuning. In future works, we would investigate the possibilities of a learnable resolution (self-loop) in the graph operator that is optimized end-to-end together with the system, essentially turning meta-learning the self-loop that guides the representation learning on graphs. Also, we would like to seek other possible theoretically-inspired approaches to alleviate training difficulties.

\bibliographystyle{abbrv}
\bibliography{references}

\end{document}